\documentclass{article}

\PassOptionsToPackage{numbers, compress}{natbib}

    \usepackage[final]{tackling_climate_workshop_style}

\usepackage[utf8]{inputenc} 
\usepackage[T1]{fontenc}    
\usepackage{hyperref}       
\usepackage{url}            
\usepackage{booktabs}       
\usepackage{amsfonts}       
\usepackage{nicefrac}       
\usepackage{microtype}      

\usepackage{custom}
\title{Attention-Based Scattering Network\\for Satellite Imagery}

\author{Jason Stock \& Charles Anderson\\
Computer Science\\
Colorado State University\\
\texttt{\{stock,anderson\}@colostate.edu}
}

\begin{document}

\maketitle

\begin{abstract}
Multi-channel satellite imagery, from stacked spectral bands or spatiotemporal data, have meaningful representations for various atmospheric properties. Combining these features in an effective manner to create a performant and trustworthy model is of utmost importance to forecasters. Neural networks show promise, yet suffer from unintuitive computations, fusion of high-level features, and may be limited by the quantity of available data. In this work, we leverage the scattering transform to extract high-level features without additional trainable parameters and introduce a separation scheme to bring attention to independent input channels. Experiments show promising results on estimating tropical cyclone intensity and predicting the occurrence of lightning from satellite imagery.
\end{abstract}

\section{Introduction}

Machine learning has received great attention in the atmospheric science community over the past couple of years. Many satellite-based applications leverage convolutional neural networks (CNN)s for tasks including but not limited to: forecasting next-hour tornado occurrences \cite{Lagerquist2020-ra}, predicting intense convection \cite{Cintineo2020-ct}, and estimating topical cyclone intensity \cite{Chen2019-pv}. These applications create input samples with stacked channel-wise features consisting of satellite imagery at different wavelengths and train a network to recognize certain patterns. While the approach is undoubtedly effective, it is not clear how these input channels are combined in the earlier layers of the network. On the other hand, a trained forecaster may intuitively look at individual channels, or the differences between multiple channels, to identify relevant features and patterns that can easily be explained. Furthermore, these networks are often limited by the quantity of available labeled data, which can lead to a model that underperforms with too few parameters or overfits as complexity increases. This further motivates the need for an architecture that is both interpretable-by-design and generalizable to small datasets. 

One effective technique to modeling sparsely labeled data is with the scattering transform as was introduced by Mallat \cite{Mallat2012-xp}. This uses a cascade of wavelet transforms with a predetermined filterbank and a non-linear modulus, akin to the structure of CNNs. Not only has the scattering transform shown promise for applications with relatively few training samples \cite{Bruna2013-wi, Sifre2013-wk, Oyallon2019-qy, Bargsten2021-eq}, but it also has many nice properties for modeling satellite data. Specifically, the design builds strong geometric invariants (e.g., translations, rotations, and scaling) and is stable to the action of diffeomorphisms -- a desirable trait due to the continuous change in cloud structure over time. Studies have also shown the scattering transform to promote sparse representation of data with a high degree of discriminability which can ultimately simplify downstream tasks \cite{Bruna2013-wi, Mallat2016-qt}. 

To build an architecture that more closely aligns with the visual interpretations of satellite imagery done by forecasters, we incorporate attention into the early layers preceding the scattering transform. Attention mechanisms work to identify salient regions in complex scenes as inspired by aspects of the human visual system \cite{itti1998model}. Recent computer vision studies have shown attention to increase performance and interpretability as well as improve confidence of post hoc explainability methods \cite{woo2018cbam, Song_undated-ot}. Most similar to this work are the studies from \cite{Zeng2020-fd, Bargsten2021-eq}. In \cite{Zeng2020-fd}, residual layers mix the input channels before applying attention and \cite{Bargsten2021-eq} applies a scattering attention module after each step in a U-Net. However, our approach differs in that we introduce a separation scheme that applies attention to individual input channels that directly follow the scattering transform.

\section{Methodology} \label{sec:methodology}

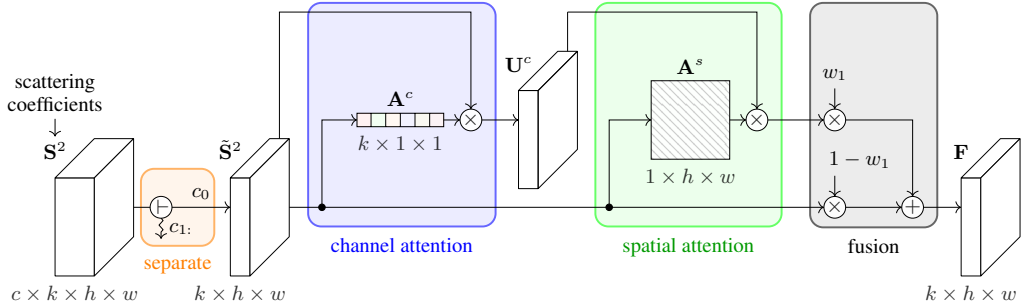
\begin{figure*}[!t]
    \centering
    \scalebox{0.87}{
\begin{tikzpicture}[
	font={\footnotesize},
	node distance=.25,
	ovr/.style={fill=white,fill opacity=.9},
	ten/.style={draw,ovr},
	ops/.style={op,ovr},
	rec/.style args={(#1/#2)}{draw,rectc,minimum width=#1cm,minimum height=#2cm,preaction={ovr}},
	recc/.style args={(#1/#2/#3)}{scale=#3,draw,rectangle split,rectangle split horizontal,rectangle split parts=#1,preaction={ovr}},
	dim/.style={text opacity=.8,inner sep=2pt,below=2pt of #1},
	sym/.style={above=3pt of #1},
	symt/.style={above=8pt of #1-northwest},
	back/.style={draw=#1!60,fill=#1!30,fill opacity=.25, line width=0.3mm,rounded corners=2mm,inner sep=4},
	xback/.style={back=#1,inner xsep=6},
	yback/.style={back=#1,inner ysep=12},
]

\matrix[
	tight,
	row sep={38,between origins},column sep=12,
	cells={scale=.3,},
	nodes={node distance=.25},
] {
    \&[-2ex]\&[-1ex]\&\&
    \node[recc=(6/1/1.3), rectangle split part fill={red!50, green!50, orange!50, blue!50, olive!50, purple!50}] (lc) {};
	\node[sym=lc] {$\vA^c$};
	\node[dim=lc] {$k \times 1 \times 1$};
    \node[ops,right=of lc] (m1) {$\times$};
    \&[-0.5ex]
	\para[draw]{mid}{0,-2.5,-2}{1,5,4};
	\node[symt=mid] {$\vU^c$};
    \&\&
    \node[rec=(4/4), pattern=north west lines, pattern color=black!20] (ls) {};
	\node[sym=ls] {$\vA^s$};
	\node[dim=ls] {$1 \times h \times w$};
    \node[ops,right=of ls] (m2) {$\times$};
    \&\&
    \node[ops] (mw1) {$\times$};
    \&\&\&[-3ex]
    \\
	\para[ten]{if}{0,-2.5,-2}{2,5,4};
	\node[ovr,dim=if-south] {$c \times k \times h \times w$};
	\node[symt=if] (scattering) {$\vS^2$};
	\&
	\node[ops] (s0) {$\,\vdash$};
	\&
	\para[ten]{bn}{0,-2.5,-2}{1,5,4};
	\node[ovr,dim=bn-south] {$k \times h \times w$};
	\node[symt=bn] {$\tilde{\vS}^2$};
	\&
	\node[dot](s1) at(0,0){};
	\&\&\&
	\node[dot](s2) at(0,0){};
	\&\&\&
	\node[ops] (mw2) {$\times$};
	\&\&
	\node[ops] (f) {$+$};
	\&
	\para[ten]{of}{0,-2.5,-2}{1,5,4};
	\node[dim=of-south] {$k \times h \times w$};
	\node[symt=of] {$\vF$};
    \\
};

\node[above=2ex of mw1] (w1) {$w_1$};
\draw[->] 
    (w1) edge (mw1)
    node[above=2ex of mw2] {$\qquad\:\, 1 - w_1$} edge (mw2)
    (mw1) -| (f)
    ;

\node[below=2ex of s0] (sep) {}; 
\draw[->, line join=round,
    decorate, decoration={
    zigzag,
    segment length=4,
    amplitude=.9,post=lineto,
    post length=2pt
}]  (s0) -- node[right] {$c_{1:}$} (sep);

\draw[->] (s0) edge  node[above] {$c_0$} (bn-west);
\draw[->] 
    (if-east) -- (s0)
    (bn-east) edge (mw2)
    (mw2) edge (f)
    (f) -- (of-west)
    ;
    
\draw[->] (s1) |- (lc);
\draw[->] 
    (lc) edge (m1)
    (m1) -- (mid-west)
    ;

\draw[->] (s2) |- (ls);
\draw[->] 
    (ls) edge (m2)
    (m2) -- (mw1)
    ;

\draw (mid-north) -- +(0,.6) coordinate (mid-n);
\draw[->] (mid-n) -| (m2);

\path (s1 |- mid-n) coordinate (s-n);
\draw[->] (bn-north) |- (s-n) -| (m1);

\coordinate[right=3.2ex of s0] (s0pad);
\coordinate[below=0.65ex of s1] (s1pad);
\coordinate[below=0.65ex of s2] (s2pad);

\begin{pgfonlayer}{bg2}
    \node[yback=orange,fit=(s0) (s0pad)] (separate) {};
	\node[xback=blue,fit=(s-n) (s1pad) (m1)] (channel) {};
	\node[xback=green,fit=(mid-n -| ls) (s2pad) (m2)] (spatial) {};
	\node[xback=black,fit=(w1 |- mid-n) (mw2) (f)] (fusion) {};
\end{pgfonlayer}

\node[black,above=10pt of scattering] (scl) {scattering\\coefficients};
\draw[->] (scl) -- +(0,-0.68) coordinate;

\node[orange,below=1pt of separate]{separate};
\node[blue,below=1pt of channel]{channel attention};
\node[green!60!black,below=1pt of spatial]{spatial attention};
\node[black,below=1pt of fusion]{fusion};

\end{tikzpicture}
}
    \caption{Network architecture illustrating the separation of attention modules on the scattering transform. The left most block represents the output of the scattering transform on the input. The \textcolor{orange}{separate} operator isolates a single channel, e.g., $C_0$, and passes the normalized scattering coefficients, $\tilde{\vS}^2$, through \textcolor{blue}{channel attention} and \textcolor{green!60!black}{spatial attention} before \textcolor{black}{fusion}. There are $C$ total attention modules in the network. Figure modified from \cite{Song_undated-ot}.}
    \label{fig:network}
\end{figure*}

\fref{fig:network} illustrates the primary components of our network, starting with our output of the scattering transform and showing an attention module separated by input channel. The implementation and design choice for each part is described in detail below.

\paragraph{Scattering Transform} Scattering representations yield invariant, stable (to noise and deformations), and informative signal descriptors with cascading wavelet decomposition using a non-linear modulus followed by spatial averaging. Using the Kymatio package \cite{JMLR:v21:19-047}, we compute a 2D transform with a predetermined filter bank of Morlet wavelets at $J=3$ scales and $L=6$ orientations. For each input channel, we apply a second-order transform to obtain the scattering coefficients $\vS^2$. These channels are processed independently and combined later in the network. Additional details on the scattering transform can be found in \aref{app:scattering-transform}.

\paragraph{Channel Separation} Local attention methods routinely process their input using all the channel information at once, e.g., feature maps from RGB color channels. However, the result of the scattering transform yields a 5-dimensional tensor, $\vS^2$, where each channel, $C$, in the input has their own set of $K$ scattering coefficients. Rather than stacking the result and passing them all through the subsequent layers together, we propose to first separate the input channels and process the coefficients individually. This creates $C$ new attention modules, each with independent weights, that are processed in parallel. By following this separation scheme we add the benefit of localizing patterns in the input before joining high-level features. Thus, the interpretation of attention over individual input channels is improved significantly, especially if the channels have different meaning, e.g., temporal, visible, infrared, derived products, etc. 

\paragraph{Attention Modules} The attention modules encompass three primary components, namely:
\begin{enumerate*}[label=(\roman*)]
  \item channel attention,
  \item spatial attention and
  \item feature fusion.
\end{enumerate*}
The channel attention features are used to inform the spatial attention module before fusion via feature recalibration. Specifically, the network learns to use the spatial information over the $K$ channels to selectively highlight the more informative coefficients from the less useful ones. Not only does this offer a performance improvement to our network, but it also adds an additional layer of interpretability with channels corresponding to particular coefficients. The spatial attention features highlight the salient features in the spatial resolution of independent input channels. This differs from most computer vision problems with RGB imagery that only have one heat map for the full image. As such, our network provides a more transparent interpretation of how the spatial information in each input channel is used to form a prediction. Implementation details of each component can be found in appendices \ref{app:channel-attn}, \ref{app:spatial-attn}, and \ref{app:feature-fusion}.

\paragraph{Combining Features} The result of applying attention to the scattering coefficients of each input channel yields $C$ output filters, $\vF$, that are stacked to $\vU^f \in \mathbb{R}^{C \times K \times W \times H}$. Following could be any task specific transformation, e.g., additional convolutions, upsampling, residual connections, etc., but for our tasks we show how to design a regression and classification head to have relatively few trainable parameters. Specifically, we reshape $\vU^f$ to have $C \cdot K$ channels, which we reduce to $16$ via a pointwise convolution. This effectively combines the high-level features of each input channel. The feature maps are flattened and input to a layer with $8$ fully-connected units before a single linear output. After the convolutional and fully-connected layers are a ReLU activation for added non-linearity. 

\begin{table}[h]
    \centering
    \caption{Experimental results using $n$ training samples and $p$ parameters.}
    \label{tab:results}
    \scalebox{0.95}{
    \begin{tabular}{ c | r@{ }l r@{ }l r@{ }l r@{ }l} \toprule
        
        \multicolumn{1}{c}{} & \multicolumn{2}{c}{Scattering} & \multicolumn{2}{c}{ResNet18} & \multicolumn{2}{c}{MobileNetV3} & \multicolumn{2}{c}{Conv.} \\ 
        \multicolumn{1}{c}{$n\downarrow\;p\rightarrow$} & \multicolumn{2}{c}{(51.8K)} & \multicolumn{2}{c}{(11.2M)} & \multicolumn{2}{c}{(1.5M)} & \multicolumn{2}{c}{(268.2K)} \\ 
        \midrule[\heavyrulewidth]
        \multicolumn{9}{l}{TC Intensity, $\text{rmse}\;(\text{R}^2)$}\\
        \midrule
        1000  & \bftab 15.83 &\bftab(0.59) & 16.47 &(0.56) & 56.85 &(-4.28) & 17.51 &(0.50) \\
        5000  & \bftab 12.01 &\bftab(0.76) & 14.30 &(0.67) & 55.18 &(-3.97) & 13.34 &(0.71) \\
        10000 & \bftab 10.98 &\bftab(0.80) & 11.85 &(0.77) & 21.13 &(0.27)  & 13.81 &(0.69) \\
        30000 & \bftab 10.35 &\bftab(0.83) & 10.74 &(0.81) & 13.07 &(0.72)  & 11.68 &(0.78) \\
        47904 & \bftab 9.34  &\bftab(0.86) & 10.66 &(0.81) & 11.90 &(0.77)  & 11.67 &(0.78) \\
        \midrule[\heavyrulewidth]
        \multicolumn{9}{l}{Lightning Occurrence, $\text{acc.}\;(\text{F1})$}\\ 
        \midrule
        1000   & \bftab 86.04 &\bftab(0.85) & 73.68 &(0.74) & 62.46 &(0.39) & 78.27 &(0.74) \\
        5000   & \bftab 88.01 &\bftab(0.87) & 87.59 &(0.87) & 68.82 &(0.55) & 82.35 &(0.82) \\
        10000  & \bftab 88.87 &\bftab(0.88) & 86.33 &(0.85) & 81.46 &(0.83) & 84.37 &(0.84) \\
        50000  & \bftab 89.58 &\bftab(0.89) & 89.20 &(0.88) & 87.49 &(0.87) & 87.99 &(0.87) \\ 
        212604 & 90.46 &(0.90) & \bftab 90.51 &\bftab(0.90) & 86.87 &(0.88) & 89.57 &(0.89) \\ 
        \bottomrule
    \end{tabular}
    }
\end{table}

\section{Experiments} \label{sec:experiments}
We demonstrate the performance of our network on two separate datasets, namely estimating wind speeds from tropical storms and predicting the occurrence of lightning over previous observations. Note that the experiments serve as an outline that could extend to other tasks that leverage multi-channel inputs.

For each experiment we compare results to a handcrafted CNN (named Conv) inspired by \cite{Maskey2020-zw}: three convolutional layers with $8, 16, \text{and } 32$ filters, each followed by ReLU and max pooling before a fully-connected layer with $32$ units and a linear output unit. Further inspiration is taken from (a subset of) recent state-of-the-art vision models, namely ResNet18 \cite{he2016deep} and MobileNetV3 (small) \cite{howard2019searching}, to better understand how larger and more complex networks compare with our proposed method.

\subsection{Estimating Tropical Cyclone Intensity}
Tropical cyclones are among the most devastating natural disasters, causing billions of dollars of damage and significant loss of life every year. Predicting the track of these cyclones is well studied, but there is still an imperative need to improve upon the forecast of intensity \cite{cangialosi2020recent}. The NASA Tropical Storm Wind Speed Competition \cite{tcdatasetv1} was released to study new automated and reliable methods of forecasting intensity. The data are single-band infrared images (i.e., band-13 or \SI{10.3}{\micro\metre}) captured by the Geostationary Operational Environmental Satellite (GOES)-16 Advanced Baseline Imager (ABI), with pixel values representing heat energy in the infrared spectrum, normalized to grayscale. We leverage the temporal relationships of previous timesteps up to the point of prediction to estimate the maximum sustained surface wind speed. Additional details can be found in \aref{app:tropical-cyclones}.

The state-of-the-art reaches a root-mean-squared error (RMSE) of \SI{6.256}{\knot} with an ensemble of 51 models \cite{ivanov1st}. We omit a direct comparison as interpreting these models would be increasingly difficult for end users. The proposed scattering network, with significantly fewer parameters, performs best overall with a minimum RMSE of \SI{9.342}{\knot} when using all available data for training. This is $12.35\%$ lower than the closest competitor, ResNet18, and $21.44\%$ and $19.92\%$ lower than MobileNetV3 and Conv, respectively (\tref{tab:results}). As such, the competing networks are more prone to overfit or lack the complexity to generalize, especially as the training size, $n$, decreases and for high-wind events (see \fref{fig:regression_summary} for additional comparisons). By leveraging the high-level features from the scattering coefficients, we maintain competitive performance even with $n=5000$ training samples. 

The local attention features $\vA^c$ and $\vA^s$ are visualized for each input channel to reveal additional insights to the network's prediction (details on how figures are computed can be found in \aref{app:feature-vis}). The example shown in \fref{fig:tc-attn} displays the spatial attention, $\vA^s$, with structural highlights having greater weight generally in the center near the storm's eyewall, where the strongest winds are usually found. Interestingly, we see points with the greatest weight along the edge of the inner rainband at timesteps $t-18$ and $t-9$, and at the eyewall at $t$. Regions with lower attention values are commonly found in the environment between rainbands. Channel attention, $\vA^c$, generally shows an increase in weight as the structure of the cyclone intensifies. We speculate this to correspond with the coefficients that are strongest along the direction of edges in the imagery. This can be observed with timestep $t-18$ having lower first-order features and considerably less variability than $t$ at different indices.

\subsection{Short Range Lightning Prediction}

Accurate short-term prediction of lightning onset can help protect life and mitigate the economic impacts from disrupted outdoor work and natural fires by updating people on when to seek shelter and the persistence of lightning events. The AI for Earth System Science Hackathon \cite{ai4es2020} opens this challenge with data from GOES-16 ABI and aggregate lightning flash counts, lagged by one hour, from the Geostationary Lightning Mapper (GLM). The input channels include the following four water vapor bands: upper-level troposphere (band-8 or \SI{6.2}{\micro\metre}), mid-level troposphere (band-9 or \SI{6.9}{\micro\metre}), low-level troposphere (band-10 or \SI{7.3}{\micro\metre}), and longwave (band-14 or \SI{11.2}{\micro\metre}). The target flash counts are converted to binary labels and used for predicting if lightning is present over the previous hour. Additional dataset details can be found in \aref{app:lightning}.

To the best of our knowledge there are no public benchmarks of this dataset. Thus, we conduct experiments with only the aforementioned models. The effect of sample size on the scattering network is minimal with $4.42\%$ lower classification accuracy when $0.47\%$ of the data is used for training. By contrast, ResNet18, with $0.05\%$ higher overall accuracy compared to the scattering network, has a $12.36\%$ lower accuracy when the same $n=1000$ samples are used for training (\tref{tab:results}). Accuracy is also $7.77\%$ higher with the scattering network over the CNN using the same training samples. \fref{fig:li-attn} displays the attention features for a given example. The spatial feature map, $\vA^s$, shows that for bands 8, 10, and 14 there is higher weight on the convective cells, where lower brightness temperatures are present. Band 9 displays a relative inverse of weight with higher values surrounding the convective cell. The channel features, $\vA^s$, show that the scenes with a convective cell have fewer first-order coefficients with higher weight. Specifically, there are dominant large scale wavelets, at a single orientation, that are more important for the network's prediction.

\section{Conclusion} \label{sec:conclusion}

In this work we introduce an attention-based scattering network that can serve as the early layers of a neural network. Our proposed separation scheme defines the most salient features and scattering coefficients on individual input channels and can easily be visualized to better understand the use of each channel. The result is a network that promotes interpretability and can easily be adapted to other satellite-based tasks. Findings show that our network, even with fewer trainable parameters than a linear model, achieves $\sim20\%$ lower error with better generalization than a standard CNN and state-of-the-art vision models on a sample application of estimating tropical cyclone intensity. For this dataset the network is effective under all constraints where data is limited. Our model shows similar results on the short-term predictions of lightning occurrence with great advantages primarily for small sample sizes, while there are diminishing returns for very large sample sizes.

In future work we seek to better understand the individual components of our network and to improve upon the fusion of individual channels. By performing an ablation study we could identify which attention features contribute most to the improved performance and have a more robust understanding of computations. It would also be useful to evaluate other methods of combining high-level features, such as an informed method for selecting the top-$n$ weighted scattering coefficients after each attention module or simpler aggregate functions. Lastly, as this method could be extended to different tasks, we would like to explore the results of applying our method to other satellite-based applications. 

\acksection
This work is supported by NSF Grant No. 2019758, \textit{AI Institute for Research on Trustworthy AI in Weather, Climate, and Coastal Oceanography (AI2ES)}.

\bibliography{main}

\newpage
\appendix
\section*{Appendix}

\section{Implementation Details} \label{app:implementation}

\subsection{Scattering Representations} \label{app:scattering-transform}

The wavelet transform of a 2D signal, $x(u)$ with $u$ denoting the spatial index, at scale J is defined as
\begin{equation}
        \mathcal{W}_J x(u) \coloneqq \big\{x * \phi_{2^J}(u)\;,\; x * \psi_\lambda(u)\big\}_{\lambda \in \Lambda_J},
\end{equation}
where $\psi_\lambda(u) = 2^{-2j} \psi(2^{-j}r^{-1} u)$ with $\lambda = 2^j r$ for $0 \leq j < J$ and $r \in G^+$ as the discrete, finite rotation group of $\mathbb{R}^2$ with $L$ equally spaced angles from $[0, \pi)$. The traditional mother wavelet, $\psi(u)$, in the scattering transform is the Morlet wavelet with a scaled Gaussian lowpass filter, $\phi_{2^J}(u) = 2^{-2J}\phi(2^{-J} u)$. A wavelet transform is translation covariant, and thus invariance measures are extracted by computing a non-linear complex modulus, $| x + iy | = \sqrt{x^2 + y^2}$, and averaging the result. The information lost during averaging is restored by applying a new wavelet decomposition with scales $j_1 < j_2$, producing new invariants. 

By following an iterative scheme we can compute $m$-th order coefficients, although here we compute up to the second order as higher orders have negligible energy \cite{Bruna2013-wi}. The zeroth-order coefficient is computed as $S^0 x(u) = x * \phi_{2^J}(u)$ and downsampled by a factor of $2^J$. To recover the high-frequency information, we perform our first wavelet transform, apply a non-linear modulus, and average again. Formally, the first-order coefficients are found by
\begin{equation}
        S^1 x(\lambda_1, u) \coloneqq |x * \psi_{\lambda_1}| * \phi_{2^J}(u).
\end{equation}

The resulting feature maps have the same resolution as $S^0$ but with $J L$ channels. Second-order coefficients are computed similarly on $S^1$ using all rotations but for smaller coefficients, denoted by
\begin{equation}
        S^2 x(\lambda_1, \lambda_2, u) \coloneqq ||x * \psi_{\lambda_1}| * \psi_{\lambda_2}| * \phi_{2^J}(u),
\end{equation}
which results in feature maps with $\frac{1}{2} J (J - 1) L^2$ output channels. Thus, if we assume $x$ to be a tensor of size $(B,C,W,H)$, then the output via a second-order scattering transform, $\vS^2$, with scale $J$ and $L$ angles will have size $(B, C, 1 + L J + \frac{1}{2} J (J - 1) L^2, W / 2^J, H / 2^J)$. Note, for brevity, in the following subsections we use $W \times H$ to denote the spatial dimension that actually occur over $W / 2^J \times H / 2^J$. 

\subsection{Channel Attention} \label{app:channel-attn}

Following the design of the squeeze and excitation block in \cite{Hu2020-my}, this attention module emphasises the local channel information of individual inputs. We use $\tilde{\vS}^2 = [\vs_1,\vs_2,\dots,\vs_K]$ as the normalized coefficients from the separated input channel of $\vS^2$. First, we \textit{squeeze} $\vs_k$ to obtain a global information embedding $\vz \in \mathbb{R}^K$ via global average pooling, where the $k$-th element, $z_k$, is
\begin{equation}
        z_k \coloneqq \frac{1}{H W} \sum_{i=1}^{H} \sum_{j=1}^{W} {s_k(i,j)}.
\end{equation}

Thereafter, we aggregate the the embedding with the \textit{excitation} operation to adaptively recalibrate channel-wise features. This results in our \textit{channel attention weights} $\vA^c \in \mathbb{R}^K$ with scalar elements, $a_k$, computed as
\begin{equation}
       \vA^c \coloneqq \sigma(g(\vz; \vv, \vw)) = \sigma(\delta(\vz \vv) \vw ),
\end{equation}
where $\sigma$ and $\delta$ refer to the sigmoid and ReLU functions, $\vv \in \mathbb{R}^{K \times \frac{K}{r}}$, and $\vw \in \mathbb{R}^{\frac{K}{r} \times K}$. Weight matrices $\vv$ and $\vw$ are initialized without bias parameters and use a reduction ratio of $r = 16$. This ratio value was shown by \cite{Hu2020-my} to be a sufficient starting parameter across multiple experiments. The final output of the channel attention weighs the normalized coefficients with the scalar elements $a_k$ to get $\vU^c \in \mathbb{R}^{K \times H \times W}$, where the $k$-th filter, $\vu_k$, is
\begin{equation}
    \vu_k \coloneqq a_k \cdot \vs_k.
\end{equation}

\subsection{Spatial Attention} \label{app:spatial-attn}

We implement a spatial attention module that predicts the importance of regions within the scattering coefficients based on the image context, similar to \cite{Jin_Kim_undated-ij, Song_undated-ot}. We first apply a pointwise convolution to our normalized coefficients with $r \times 1 \times 1$ filters, using a reduction ratio of $r=16$, to compress channel dimensionality. Spatial context is then extracted from the resulting feature map by convolving three dilated convolutions of size $3 \times 3$ with dilation factors of one, two, and three. These dilations increase the receptive field while preserving the input resolution. This allows us to stack the four feature maps to create a $4r \times W \times H$ tensor. This is reduced via a pointwise convolution using a $1 \times 1 \times 1$ filter (per channel dimension) to yield the final \textit{spatial feature map} $\vA^s \in \mathbb{R}^{1 \times W \times H}$. 

The spatial feature map and channel weighted coefficients are multiplied together to get a complete local attention map, $\vU^s \in \mathbb{R}^{K \times W \times H}$, as given by
\begin{equation}
    \vU^s \coloneqq \vA^s \odot \vU^c,
\end{equation}
where $\odot$ is the Hadamard product. At a high level, $\vU^s$ contains all the normalized scattering coefficients where each coefficient (i.e. $k$-th channel) is weighted and spatially scaled to a localized region. This information is used downstream to highlight the positive or negative features of the scattering transform.

\subsection{Feature Fusion} \label{app:feature-fusion}

The normalized scattering coefficients and local attention maps are combined in the fusion block at the end of an attention module. We follow a method similar to \cite{tan2020efficientdet} and \cite{Song_undated-ot} to get an output from the weighted average of features. A single parameter, $w_1$, with an initial value of $0.5$ is trained to assign contributions of each pathway. Mathematically, 
\begin{equation}
    \vF \coloneqq w_1 \vU^s + (1 - w_1) \tilde{\vS}^2,
\end{equation}
where $w_1$ is clamped on the interval $[0, 1]$ after each update to ensure the contributions sum to one.

\section{Dataset Details} \label{app:dataset-details}

\subsection{Tropical Cyclones} \label{app:tropical-cyclones}

\begin{figure*}[!ht]
    \centering
    \input{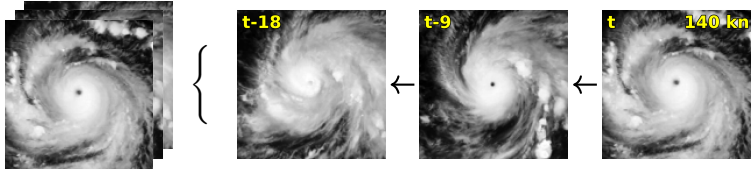}
    \caption{Input data of size $(1 \times 3 \times 128 \times 128)$ with stacked channel-wise timesteps. The target variable is the wind speed associated with the last frame.}
    \label{fig:tc_samples}
\end{figure*}

The topical cyclone dataset contains a collection of satellite imagery for over 600 storms in the Eastern Pacific and Atlantic Oceans from $2000$ to $2019$. Test data consists of imagery from storms not included in the training data as well as held out samples from later in a storm's life cycle. Since observations from temporal data are not independent, we seek to reduce the implications of autocorrelations (i.e., from trends and seasonality) and holdout imagery from the last $20\%$ of each storm to create the validation set.

To extend from the single channel imagery to multi-channel inputs we leverage the temporal relationships of previous timesteps up to the point of prediction. Three frames separated by a nine step interval (i.e., $t-18, t-9, \text{and } t$) are stacked to create a $3 \times 128 \times 128$ input sample using the last frame’s wind speed as the target value (\fref{fig:tc_samples}). Inputs are created following the next $t+1$ timestep and repeated over all datasets yielding $47,904$ training, $7,119$ validation, and $37,913$ test samples. These images are min-max normalized to the interval [0,1] to stabilize the result of the scattering transform. Target wind speeds are z-score normalized to have zero mean and unit variance using the statistics of the training data. Predicted and target values are unnormalized after inference for evaluation.

When subsampling the training data to smaller values of $n$, we define $m=11$ equally spaced boundaries, $B = \{r : r = 15 + 17k\;|\;k = m-1, m-2, \dots, 0\}$, such that $n$ total samples that comprise the entire training dataset are loosely divided into groups following $B_i \leq t > B_{i+1}$ for targets $t$. Data within each group are randomly sampled and may borrow from the $B_{i+1}$ boundary to ensure an equal distribution of target wind speeds. \tref{tab:results} defines the values of $n$ used in our experiments.

\subsection{GOES-16 Lightning} \label{app:lightning}

\begin{figure*}[!ht]
    \centering
    \input{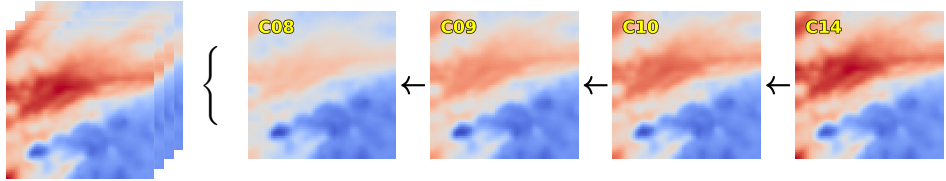}
    \caption{Input data of size $(1 \times 4 \times 64 \times 64)$ with stacked channel-wise brightness temperatures. The target variable is a binary label indicating the presence of lightning over the previous hour.}
    \label{fig:li_samples}
\end{figure*}

The GOES-16 data consists of $32 \times 32$ image patches (for each band) across the Continental United States (between latitudes \ang{29.09} and \ang{48.97} and longitudes \ang{-97.77} and \ang{-82.53}) at 20 minute intervals from $2019\text{-}03\text{-}02$ through $2019\text{-}10\text{-}01$. We perform bilinear interpolation to each band to scale inputs to the resolution of $64 \times 64$ for more accurate spatial attention features. This is done because the scattering transform yields coefficients of resolution $W / 2^J \times H / 2^J$ (i.e., $8 \times 8$) and resolutions too small will lose detail. The brightness temperatures, measured in kelvins, are min-max normalized on the interval $[0,1]$ using the statistics of the training data. In total there are $212,604$ training, $212,604$ validation, and $199,157$ test samples. \fref{fig:li_samples} displays an example input sample with each stacked channel-wise bands.

Flash counts from the GLM have a strong positively skewed distribution (i.e., $4.47 \pm 18.22$) across all training samples. When converting to binary labels, where true when flash counts are greater than zero, we get a better distribution of targets with a slight class imbalance of $63.49\%$ training samples having lightning. When subsampling the training data we reduce bias by maintaining the class distribution such that there are $n$ total samples ranging from $1,000$ to $212,604$ (\tref{tab:results}).

\section{Feature Visualizations} \label{app:feature-vis}

\subsection{Spatial Attention Features}

The separation of attention modules yield a spatial attention feature map, $\vA^s$, for each input channel in our data. This map can be visualized via a bilinear upsampling from $W / 2^J \times H / 2^J$ to our original input resolution $W \times H$. The scaled feature map can be superimposed on the original data, and the detail of this map will depend on the scale, $J$, of the scattering transform.

\subsection{Channel Attention Features}

For first-order coefficients, the polar radius is inversely proportional to scale $2^{j_1}$ for the wavelet $\psi_{\lambda_1}$ with an angle corresponding to the rotation $r_1$ (approximating the frequency bandwidth of the Fourier transform $\hat{\psi}_{\lambda_1}$). Thus, each quadrant can be indexed by $(r_1, j_1)$. Second-order coefficients are displayed similarly by subdividing each first-order quadrant, $(r_1, j_1)$, along the polar radius with $j_1 < j_2 < J$ for all $L$ angles, which can be indexed by $(r_1, r_2, j_1, j_2)$. However, instead of visualizing individual scattering coefficients, we display each scalar element in $\vA^c$ (normalized between $[0, 1]$) corresponding to these indices and input channel.

\subsection{Gradient Based Methods}

The scattering transform is differentiable, and thus, allows for evaluations of post hoc explainability methods. We demonstrate an example of how gradients can be computed with respect to an individual input pixel by computing integrated gradients for our example input. Following the work of \cite{Sundararajan2017-bb}, we establish a baseline of all zeros and compute importance scores for each pixel in the input. While we show integrated gradients as an example, alternative post hoc explainability methods, e.g., GradCAM, layer-wise relevance propagation, Shapley values, etc., could be used to evaluate this network.

\section{Supplemental Figures}

\begin{figure}[ht]
    \centering
    \begin{subfigure}[t]{0.49\linewidth}
        \centering
        \includegraphics[width=\linewidth]{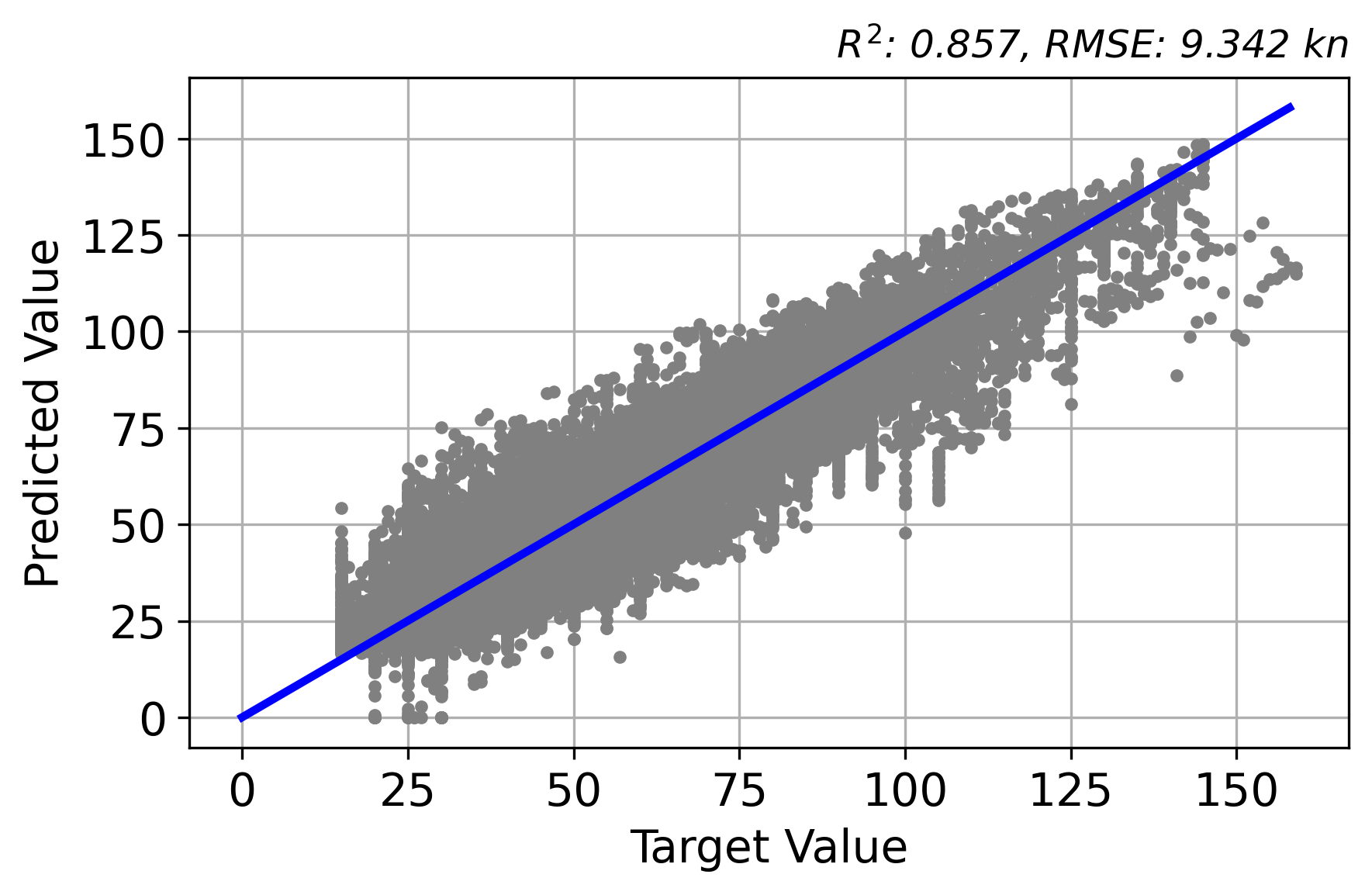}
        \caption[b]{Scattering Network}
        \label{fig:regression_summary.scattering}
    \end{subfigure}
    \hfill
    \begin{subfigure}[t]{0.49\linewidth}
        \centering
        \includegraphics[width=\linewidth]{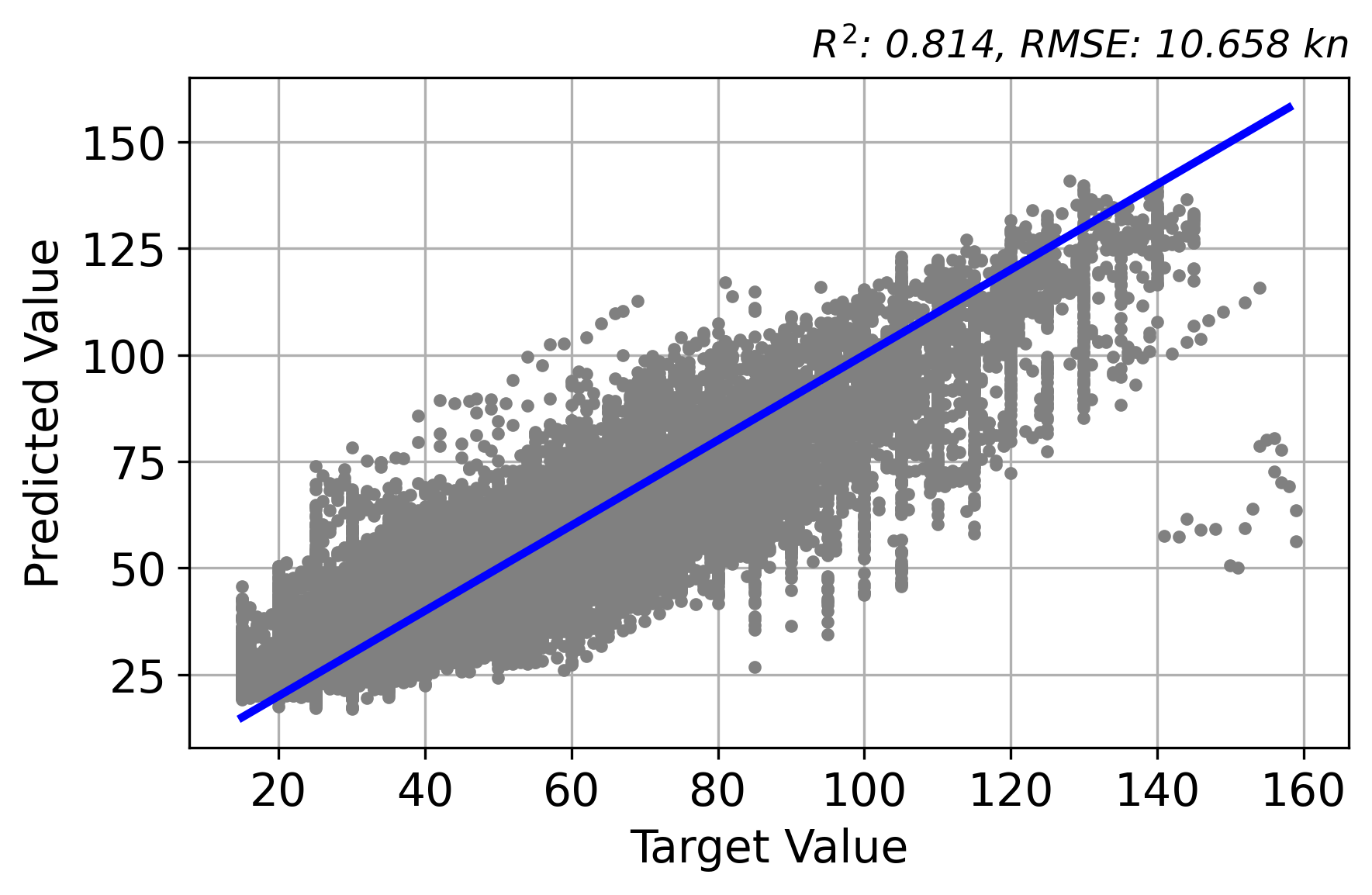}
        \caption{ResNet18}
        \label{fig:regression_summary.resnet18}
    \end{subfigure}
    \caption{Target vs. predicted wind speeds from the \textbf{(a)} proposed scattering network and \textbf{(b)} ResNet18 network, trained using all the available data. The greatest errors are observed with the highest intensity samples, and evidently, is where ResNet18 performs worst. Target wind speeds $>$\SI{140}{\knot} have an $\text{RMSE}=\;$\SI{51.630}{\knot} from ResNet18 as compared to an $\text{RMSE}=\;$\SI{27.231}{\knot} from the scattering network.}
    \label{fig:regression_summary}
\end{figure}

\begin{figure*}[!ht]
    \centering
    \input{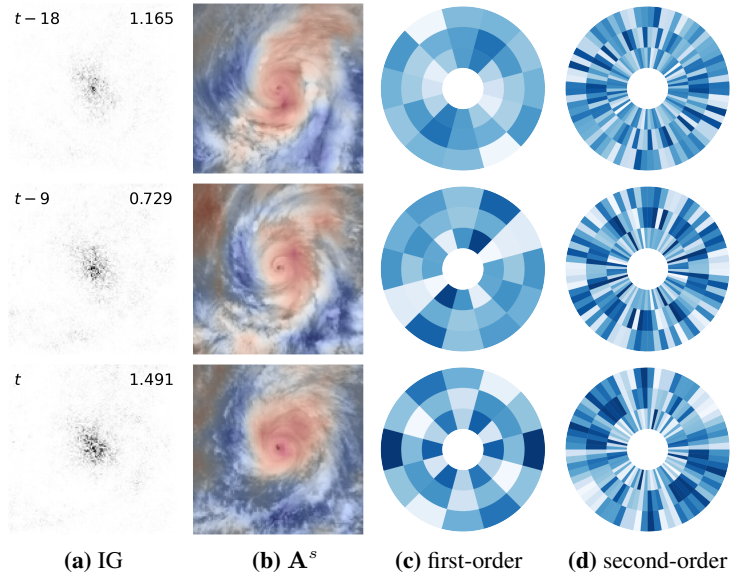}
    \caption{Feature visualizations from the tropical cyclone intensity data.}
    \label{fig:tc-attn}
\end{figure*}

\begin{figure*}[!ht]
    \centering
    \input{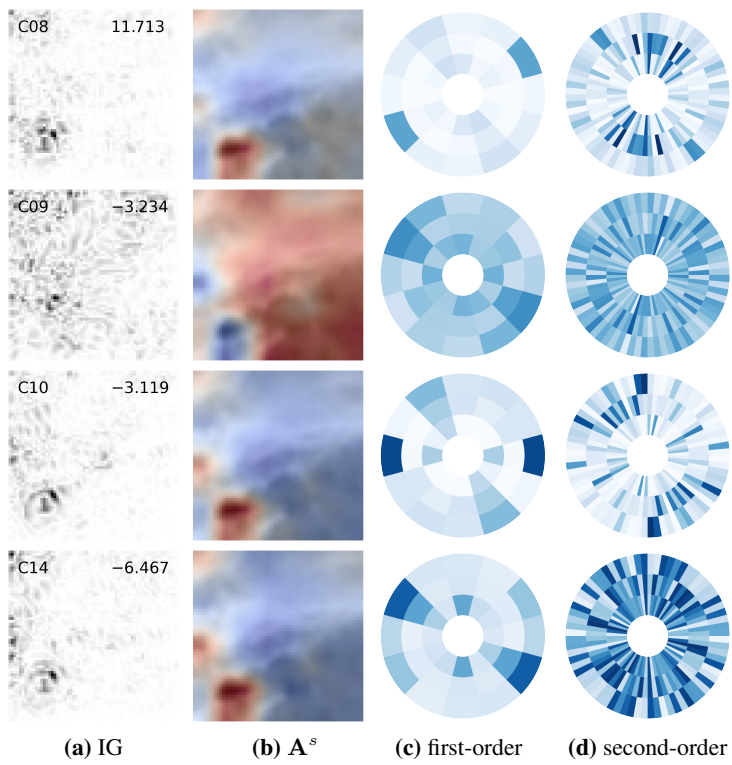}
    \caption{Feature visualizations from the short-term lightning prediction data.}
    \label{fig:li-attn}
\end{figure*}

\end{document}